# Real-Time Magnetic Tracking and Diagnosis of COVID-19 via Machine Learning

Dang Nguyen[1,2#], Phat K. Huynh[3#], Vinh Duc An Bui[4,*], Kee Young Hwang[1], Nityanand Jain[5], Chau Nguyen[6], Le Huu Nhat Minh[7], Le Van Truong[8], Xuan Thanh Nguyen[9], Dinh Hoang Nguyen[10], Le Tien Dung[11], Trung Q. Le[2,3,*], and Manh-Huong Phan[1,*]

[1] Department of Physics, University of South Florida, Tampa, FL 33620, USA

[2] Department of Medical Engineering, University of South Florida, Tampa, FL 33620, USA

[3] Department of Industrial and Management Systems Engineering, University of South Florida, Tampa, FL 33620, USA

[4] Department of Thoracic and Cardiovascular Surgery, Hue Central Hospital, Hue City, Vietnam

[5] Faculty of Medicine, Riga Stradiņš University, 16 Dzirciema street, Riga, LV-1007, Latvia

[6] Vietnam National University, Ho Chi Minh City, Vietnam

[7] Emergency Department, University Medical Center, Ho Chi Minh City, Vietnam

[8] Traditional Medicine Hospital, Ministry of Public Security, Hanoi 10000, Vietnam

[9] Department of Abdominal Emergency and Pediatric Surgery, Hue Central Hospital, Hue City, Vietnam

[10] Department of Adult Cardiovascular Surgery, University Medical Center Ho Chi Minh City, Ho Chi Minh City, Vietnam

[11] Department of Lung, University Medical Center Ho Chi Minh City, Ho Chi Minh City, Vietnam

[#] Co-first authors

[*] Co-corresponding authors: phanm@usf.edu (M.H.P.), tqle@usf.edu (T.Q.L.),

buiducanvinh@gmail.com (V.D.A.B.)

The COVID-19 pandemic underscored the importance of reliable, noninvasive diagnostic tools for robust public health interventions. In this work, we fused magnetic respiratory sensing technology (MRST) with machine learning (ML) to create a diagnostic platform for real-time tracking and diagnosis of COVID-19 and other respiratory diseases. The MRST precisely captures breathing patterns through three specific breath testing protocols: normal breath, holding breath, and deep breath. We collected breath data from both COVID-19 patients and healthy subjects in Vietnam using this platform, which then served to train and validate ML models. Our evaluation encompassed multiple ML algorithms, including support vector machines and deep learning models, assessing their ability to diagnose COVID-19. Our multi-model validation methodology ensures a thorough comparison and grants the adaptability to select the most optimal model, striking a balance between diagnostic precision with model interpretability. The findings highlight the exceptional potential of our diagnostic tool in pinpointing respiratory anomalies, achieving over 90% accuracy. This innovative sensor technology can be seamlessly integrated into healthcare settings for patient monitoring, marking a significant enhancement for the healthcare infrastructure.

# 1. INTRODUCTION

Breathing is fundamental to human quality of life, making respiratory patterns vital health indicators. Abnormalities in these patterns often hint at respiratory diseases such as chronic obstructive pulmonary disease (COPD), obstructive sleep apnea (OSA), pneumonia, cystic fibrosis, asthma, and COVID-19 [1]. Timely diagnosis of COVID-19 has become especially critical for enhancing individual well-being and bolstering public health [2]. Symptoms of COVID-19 may encompass an elevated respiratory rate, diminished tidal volume, and irregular breathing rhythms [3]. Hence, precise and prompt evaluation of respiratory patterns is essential for the diagnosis and management of COVID-19 and its variants. Although the gold standard for COVID-19 diagnosis is the polymerase chain reaction (PCR)-based method, it is constrained by time-intensive processes, logistical challenges, and an absence of continuous monitoring capabilities. Traditional respiratory monitoring techniques, like spirometry, plethysmography, and impedance pneumography, tend to be invasive and unsuitable for continuous surveillance [4, 5]. In the midst of a pandemic, scalable, non-invasive, and contactless methods based on respiratory patterns are crucial. These methods can be widely deployed in community settings, allowing for proactive management and potential early containment of outbreaks [6].

Recent advancements in noninvasive respiratory monitoring, including respiratory inductance plethysmography (RIP) [7], electrical impedance tomography (EIT) [8], and magnetic respiratory sensing technology (MRST) [9, 10], hold the promise to address these challenges, introducing innovative platforms for the diagnosis, monitoring, and management of COVID-19 [11-13]. However, RIP's calibration is intricate and user-dependent, resulting in varied measurements. It is also prone to motion artifacts and might not reliably measure tidal volume or minute ventilation in all patients. Prolonged usage of the RIP system can be uncomfortable, hindering continuous, long-term monitoring. As for EIT, its limitations stem from requiring specialized technicians for operation and interpretation. Additionally, its spatial resolution is somewhat lacking, making pinpointing changes in the lungs challenging and potentially inaccurate. Conversely, MRST is non-invasive and gauges magnetic field alterations due to respiratory movements, eliminating the need for direct skin contact. This high-resolution technology offers real-time insights into respiratory patterns, aligning with the demand for constant breathing monitoring [9, 10]. MRST's magnetic sensing principle shields it from motion artifacts, ensuring dependable performance under varied conditions, such as during COVID-19 tests. Moreover, MRST can be fashioned into a user-friendly wearable device, expanding its utility. Given these attributes, MRST emerges as a superior solution, addressing the technical and practical hurdles posed by RIP and EIT.

Machine learning (ML) algorithms have recently been harnessed to diagnose COVID-19 using a range of data sources such as chest X-rays, CT scans, and electronic health records (EHR) [14-18]. These endeavors underscore the efficacy of ML in diagnosing COVID-19, sometimes even surpassing the acumen of human experts. Moreover, ML's prowess has been extrapolated to respiratory data to diagnose and gauge the severity of ailments like asthma [19], COPD [20], obstructive sleep apnea [21, 22], and pneumonia [23] . Its precision in pinpointing distinct disease characteristics is commendable. Several ML models have been formulated to anticipate respiratory failure in critically ill patients [24] and assess the risk of COVID-19 infection from breathing rate variations [25]. The potential of automated monitoring of respiratory biomarkers and vital signs in both homes and clinical settings has been explored [26], as has the automated diagnosis of COVID-19 through respiratory audio data [27]. Methods have been pioneered for non-invasively determining respiratory rates using photoplethysmogram and electrocardiogram signals [28]. Vetted ML algorithms, such as eARDS, show potential in the early detection of acute respiratory distress syndrome (ARDS) in COVID-19 patients [29]. Additionally, innovative techniques, such as non-invasive electronic noses based on exhalation pattern recognition [30], wearable devices measuring breathing rates [31], open clinical data resources primed for deep learning-driven COVID-19 prediction [32], and radar sensor-driven breathing detection through ML [33], display promise for refined non-invasive monitoring.

Capitalizing on the capabilities of MRST and ML, we introduced an innovative concept for a diagnostic device that can not only detect COVID-19 in patients but also monitor its progression in real-time [9, 10]. We successfully incorporated MRST to examine both healthy and COVID-19-afflicted subjects, analyzing the gathered respiratory data with sophisticated ML models. Our analysis indicates that the intricate respiratory patterns and deviations can be identified with a combination of MRST's high-resolution data capture and ML algorithms, paving the way for early disease detection, including COVID-19. Such synergy facilitates preemptive healthcare, sending timely alerts about patient status shifts, thus enabling immediate and apt interventions. Comprehensive clinical and technical validation steps will be undertaken to assess the device's efficacy, user-friendliness, and influence on patient results. This innovation has the potential to revolutionize not just the detection and monitoring of COVID-19, but to also serve as a multipurpose platform for managing various respiratory conditions across diverse healthcare environments. For an in-depth understanding, we've structured the rest of this paper as follows: Section 2 delves into the methodologies employed during the design, crafting, and validation phases of the novel diagnostic tool that merges MRST and ML. Section 3 scrutinizes results from clinical examinations of healthy and COVID-19 subjects using multiple ML frameworks. In Section 4, we dissect the salient findings, discuss the potential implications and constraints, and chart out avenues for future investigations. Finally, Section 5 encapsulates the overarching influence of these discoveries on the domain.

## 2. METHODOLOGIES

### 2.1. Magnetic respiratory detection

**Fig. 1** presents a detailed overview of our research design and the schematic of our COVID-19 monitoring and diagnosis system. The process begins with the magnetic respiratory monitoring system (**Fig. 1a**). This system employs a Hall effect sensor to detect alterations in the magnetic fields created by a small permanent magnet affixed to a person's chest. These alterations arise due to respiratory movements. A comprehensive elucidation of the operational principle of the Hall effect sensor and the magnetic respiratory monitoring system can be found in the **Supplementary Information** (see **Figs. S1 and S2** and corresponding texts). This configuration guarantees non-invasive and precise tracking of a plethora of respiratory patterns.

To establish a study baseline, we adopted a breath test protocol during respiratory data acquisition (**Fig. 1b**). This protocol encompasses three distinct breathing styles: normal breathing, breath-holding, and deep breathing. The inclusion of these varied breathing techniques ensures that a broad spectrum of respiratory activities is captured, furnishing a robust dataset for subsequent analysis.

The resulting respiratory data undergoes further refinement through specialized algorithms dedicated to signal processing and feature extraction (**Fig. 1c**). This analytical phase is vital as it aids in identifying and isolating diagnostic patterns within the respiratory data. To conclude the process, the distilled features are fed into ML models (**Fig. 1d**). These models are calibrated to discern and diagnose COVID-19 based on the specified respiratory markers.

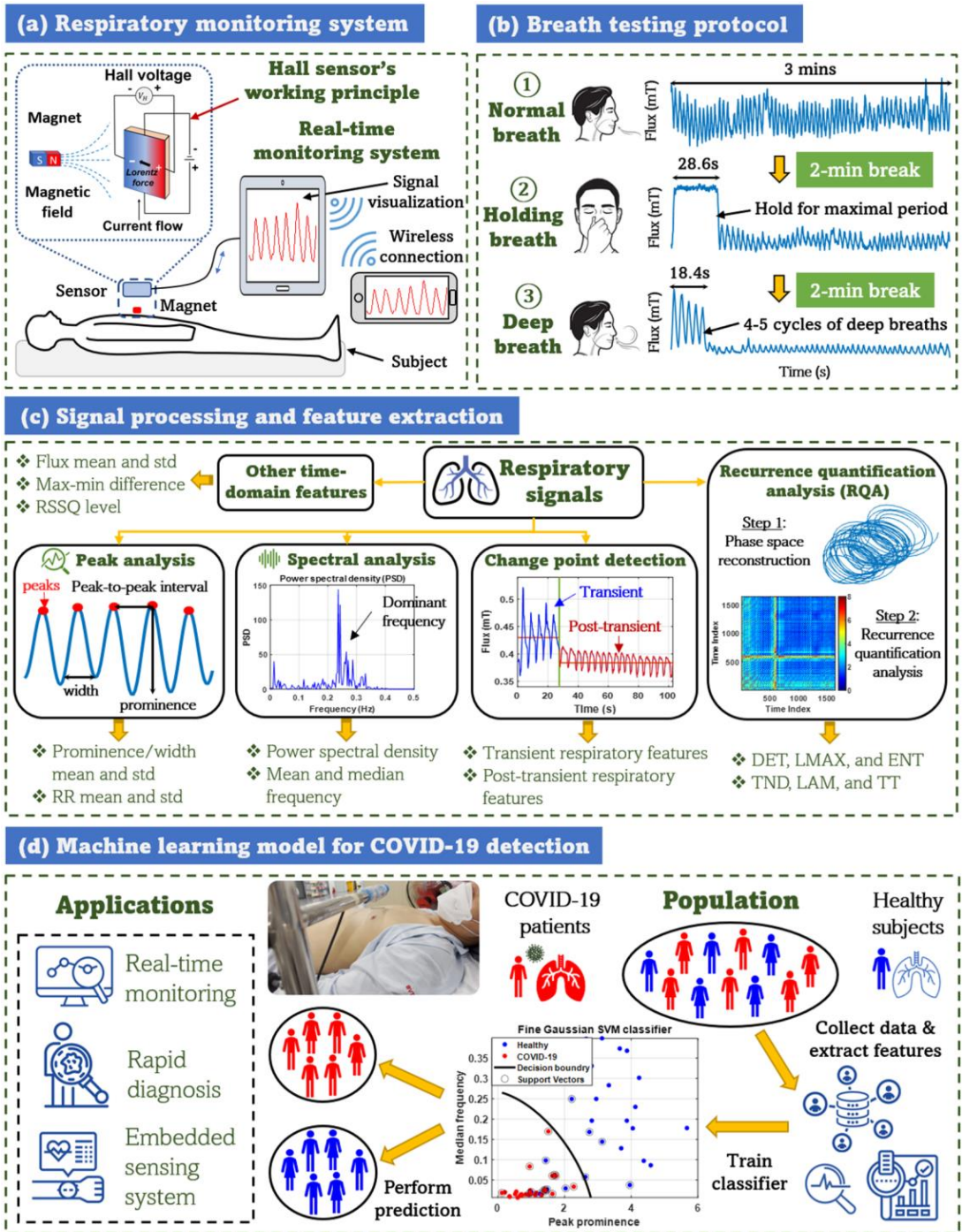

**Fig 1**. An overview of the research framework and the COVID-19 diagnosis and monitoring system. (**a**) Schematic representation of the magnetic respiratory monitoring system capturing respiratory patterns using the Hall effect sensor in tandem with a small permanent magnet; (**b**) Breath test protocol utilized during data collection, incorporating three distinct breathing styles:

normal breath, holding breath, and deep breath; (**c**) Signal processing methods and feature extraction used to analyze the acquired respiratory signals. (**d**) Illustration of ML model for COVID-19 diagnosis based on extracted respiratory features.

## 2.2. Participants and data collection

The study was carried out from July 2021 to October 2021, a period when Vietnam was enforcing the "Zero COVID" policy, necessitating compulsory centralized isolation for all COVID-19 patients either in hospitals or medical camps. Throughout the three-month duration, our project enlisted 33 COVID-19 patients from a medical camp in Binh Tan, Ho Chi Minh City, and 37 healthy participants who were residents of Ho Chi Minh City. All study subjects were selected to be over 18 years old, provided clear explanations regarding the benefits and risks of participating in the research. Only those who voluntarily chose to participate and signed a written informed consent were included. For the selection process, specific criteria were set for different groups. For the healthy participants group, the following criteria were adhered to:

1.  The participant should not have had any prior confirmation of SARS-CoV-2 infection.

2.  The participant should not have a history of either direct or indirect exposure to COVID-19 within the 14 days leading up to their participation in the study.

3.  The participant should not have been part of any COVID-19 infection clusters within 14 days of joining the study.

On the other hand, for the COVID-19 patient group, the criteria were:

1.  The patient must have a real-time PCR result with a CT index of less than 30.

2.  The patient should experience mild or no breathing difficulties.

3.  The patient's SPO2 level should be greater than 96%, regardless of whether they are breathing through an oxygen nasal cannula or just ambient air.

4. The patients should be conscious and capable of attending to their personal needs independently.

5. X-Ray shows no lung injury or injury covering less than 50% of the lung area.

Finally, certain individuals were excluded from the study based on the following criteria:

1. Those with conditions like COPD, bronchial asthma, or lung cancer.

2. Individuals with a history of pulmonary tuberculosis.

All data collection was conducted in a controlled environment, closely monitored by healthcare professionals to ensure participant safety. Before data collection, every participant was thoroughly informed about the procedures to secure their consent. To commence the test, each participant positioned a small permanent magnet on their chest, aligning with the sensor probe. They then adhered to the breath test procedure depicted in **Fig. 1b.** The protocol unfolded as follows:

1. Participants began with 3 minutes of regular breathing, allowing the MRST device to adjust and familiarize itself with the individual's typical respiratory rhythm.

2. Subsequent to this calibration, participants maintained normal breathing for another 3 minutes to gather consistent respiratory data.

3. A 2-minute respite followed, ensuring participants were comfortable and prepared for the subsequent phase.

4. Participants then attempted to hold their breath for 3 minutes. Recognizing individual variations in breath-holding capacity, we advised participants to sustain their breath as long as feasible within the designated 3-minute period.

5. After another 2-minute rest, participants engaged in 3 minutes of deep breathing, capturing heightened respiratory activity.

The inclusion of regular, withheld, and deep breathing was intentional, designed to encapsulate an extensive spectrum of respiratory behaviors, thus enriching our dataset for future analysis. The MRST's recordings, which chronicled the respiratory patterns during each test, were subsequently scrutinized to derive numerous respiratory metrics, such as time-domain features, spectral attributes, and recurrence quantification analysis (RQA) statistics [34]. Furthermore, demographic and clinical details were procured for each participant, including age, gender, height, and weight. For those diagnosed with COVID-19, we gathered supplemental data: duration of hospital stays, diastolic and systolic blood pressures, and specific COVID-19 symptoms. All respiratory data were safely stored in a HIPAA-compliant database. Each participant was allocated a distinct identifier, upholding anonymity, and privacy standards. This study was registered and received the institutional review board (IRB) approval from the ethical board of the Hue Central Hospital, under the registration number 2336/BVH.

## 2.3. Data preprocessing and feature extraction

At the data preprocessing phase, our foremost goal was to enhance the quality of the respiratory signals acquired from the MRST device. To this end, we applied a 5th order Butterworth bandpass filter tailored with a cut-off frequency (0.1–0.5 Hz) aligned to the typical human respiration frequency spectrum. This aimed to remove high-frequency disturbances, such as electromagnetic interference, and counteract the baseline wandering effect.

Subsequently, we extracted 21 pivotal features from the respiratory signal data captured by MRST. These features were then funneled into the ML models for optimized disease diagnosis. Our strategy included a blend of quantitative analysis features derived from peaks, time-domain, frequency-domain, and RQA [34] (see **Table S1**). The initial step involved peak detection in respiratory signals to capture intrinsic peak height and usage location. We gauged metrics such as mean and standard deviation of peak prominence, peak width, and the respiration rate deduced from peak-to-peak intervals. Such metrics provide insights into the intensity of breathing and the duration of each breathing cycle, with variations indicating potential respiratory anomalies, which are common symptoms in many respiratory diseases, including COVID-19.

For time-domain features, we extracted values like the maximum-to-minimum difference, root-sum-of-squares level, and the mean and standard deviation of flux amplitude. These metrics encapsulate signal amplitude characteristics, shedding light on potential abnormal respiratory patterns. On the other hand, frequency-domain features, including band power, mean and normalized power spectral density, mean frequency, and dominant frequency, elucidate the signal's periodicity and energy dispersal. Lastly, we computed RQA statistics (DET, LMAX, ENT, TND, LAM, and TT) to gain insight into the non-linear dynamic characteristics of the respiratory signals not captured by traditional time and frequency-domain analysis. The core principle of RQA is based on the construction of a recurrence plot, a graphical tool that maps recurrent events over a time series. Pertinent measures derived from this recurrence plot quantify the density, determinism, predictability, complexity, and other aspects of the underlying system dynamics of the respiratory signals. Within the context of our study, deviations in the breath cycle in respiratory signals can be viewed as a recurrent event, and abnormalities in the breath cycle due to diseases such as COVID-19 can lead to changes in RQA metrics. Collectively, these 21 features provide a multifaceted understanding of respiratory signals, bolstering the efficiency of monitoring and diagnosing COVID-19 and other respiratory diseases, while also providing additional training data to our ML model for improved diagnostic outcomes.

Beyond extracting primary respiratory features, we employed the change point detection technique [35]. This method excels in pinpointing abrupt shifts or 'change points' in respiratory data, notably emphasizing the breath-holding and deep breathing tests. The change from breath holding to normal breathing and the transient phase in between are of considerable interest, and change points typically occur around these junctures. Similarly, during the deep breathing test, participants performed 4-5 cycles of deep breathing then returned to normal breathing. Similarly, change points are expected to occur around the transition between deep breathing and normal breathing. In each case, by applying change point detection, we derived the secondary features before and after the change point (described in **Table S2**) identifying the critical transients in the respiratory pattern. By leveraging this method, we could distinguish vital transients in respiratory patterns, shedding light on physiological responses potentially impacted by COVID-19, which might otherwise remain unnoticed in regular breathing. This extensive feature set enables a more comprehensive analysis of respiratory signals, thus improving the predictive performance of our ML models.

Post feature extraction and feature selection was the subsequent imperative. Given the expansive dimension of the feature spectrum, we deployed the two-sample Kolmogorov-Smirnov (KS) test [36]. This non-parametric tool is apt for our context as it refrains from assuming any fixed distribution for the data—a valuable trait, especially considering the intricate and potentially non-Gaussian nature of respiratory signals. Through the KS test, a statistic emerges, measuring the divergence between the cumulative distribution functions (CDFs) of two feature distributions for our primary classes: healthy individuals and COVID-19 patients. A pronounced KS statistic denotes the feature's aptness in differentiating the two classes, thus such features were given precedence in the succeeding machine-learning model.

### 2.4. Machine learning models

We tailored our ML approach to construct and validate distinct models for three separate breath tests: normal, hold, and deep breathing. Each breath test utilized the pertinent set of features, extracted from its associated respiratory signals, to educate its model. Such a method aims to pinpoint disease-specific indicators unique to each test, thereby offering a holistic diagnosis.

During the training phase, data was split into two groups: training (80%) and validation (20%). This 80-20 split adheres to prevalent ML conventions. The training set, consisting of features chosen in the preceding phase, was utilized for model training, while the latter was reserved for gauging model performance on unfamiliar data. The extracted features from each breathing test served as the data input for the ML models. We opted for the Classification Learner Toolbox in MATLAB [37], due to its expansive toolkit and user-friendliness, facilitating automated model training across a broad array of classifiers including decision trees, support vector machines, logistic regression, and neural networks. By trialing these classifiers, we aimed to pinpoint the most apt model for every breath test. To bolster model robustness and applicability, 5-fold cross-validation was executed on the training data. Post training, models were evaluated using an array of metrics, such as sensitivity, specificity, accuracy, and the receiver operating characteristic (ROC) curve [38]. The ROC curve depicts the balance between sensitivity and specificity across varying thresholds, which aids in pinpointing the ideal threshold. By establishing three distinct model sets, each fine-tuned to a specific breath test, our goal is to enhance our method's diagnostic precision, capturing the intricacies inherent in each breath type, and delivering a more granular insight into a patient's respiratory health.

## 2.5. Causal analysis

Causal analysis is pivotal in deciphering the link between respiratory patterns and COVID-19 presence. We utilized the matching method [39] to minimize bias when estimating the causal impacts of COVID-19 on respiratory patterns in our study. The initial phase entailed categorizing two sets: patients diagnosed with COVID-19 and a control group of healthy individuals. Subsequently, every COVID-19 patient was paired with a healthy counterpart based on shared confounders via the optimal matching algorithm [40]. Confounders, which include attributes like age, gender, and body mass index (BMI), can potentially influence both respiratory patterns and COVID-19 statuses. Post matching, we can compare the respiratory patterns of the COVID-19 group and the control group. Using these paired sets, a paired t-test was employed to gauge the average causal effect (ACE) [41], denoting the mean outcome difference between COVID-19 patients and the controls. The ACE, as determined by this method, facilitates evaluating the significance of discerned causal impacts on continuous respiratory features. To further enhance our analysis, ACE was converted to average causal effect percentage (ACEP). For ACEP calculation, ACE was divided by the mean respiratory feature value observed in the healthy control group. This average is derived from the collective mean values of a particular respiratory trait across all healthy control participants. Consequently, this metric provides a clear understanding of the divergence in respiratory patterns between COVID-19 patients and healthy individuals.

## 3. RESULTS

### 3.1. Overview of the collected data

Our study meticulously gathered respiratory signal data from 70 participants, comprising 33 COVID-19 patients and 37 healthy individuals. MRST was employed during three breathing test types: normal, breath-holding, and deep breathing. A more detailed summary of the participants' characteristics is given in **Table 1**.

**Table 1**. Characteristics of COVID-19 patients and healthy subjects in the dataset.

| | COVID-19 patients | Healthy subjects |
|---|---|---|
| **Participants**, *n* | 33 | 37 |
| **Age**, *mean (std), years* | 46.88 (13.64) | 40.91 (16.88) |
| **Gender** | | |
|   Male, *n (%)* | 14 (42.42) | 13 (35.14) |
|   Female, *n (%)* | 19 (57.58) | 24 (64.86) |
| **Height**, *mean (std), cm* | 163.38 (8.59) | 159.77 (7.14) |
| **Weight**, *mean (std), kg* | 61.57 (10.22) | 56.38 (10.67) |
| **BMI**, *mean (std), kg/m$^2$* | 22.93 (2.99) | 22.03 (3.32) |
| **Hospitalization days**, *mean (std), days* | 5.91 (4.83) | |
| **Temperature**, *mean (std), °C* | 37.05 (0.29) | |
| **Diastolic blood pressure**, *mean (std), mmHg* | 80.87 (9.85) | |
| **Systolic blood pressure**, *mean (std), mmHg* | 136.23 (24.55) | |
| **COVID-19 symptoms** | | |
|   Difficult breathing, *n, (%)* | 16 (48.48) | |
|   Cough, *n (%)* | 20 (60.61) | |
|   Fever, *n (%)* | 15 (45.45) | |
|   Fatigue, *n (%)* | 9 (27.27) | |
|   Muscle aches, *n (%)* | 6 (18.18) | |
|   Loss of taste, *n (%)* | 3 (9.09) | |
|   Sore throat, *n (%)* | 3 (9.09) | |
|   Running nose, *n (%)* | 2 (6.06) | |

Insights from **Table 1** reveal that COVID-19 patients had, on average, a slightly older age than healthy subjects. However, the notable standard deviations underscore a wide age range in both groups, underscoring the dataset's diversity. Both groups also displayed nearly balanced gender distribution, minimizing the potential for gender-centric biases in our findings. Noteworthy is the fact that the COVID-19 patients had an average hospital stay of approximately six days, shedding light on the severity of the illness in this group, which could manifest in their respiratory patterns. Inclusion of physiological parameters such as temperature, blood pressure, and BMI, along with the noted COVID-19 symptoms, offers a comprehensive snapshot of participants' health conditions. The respiratory data sourced from the healthy participants play a crucial role as a control group. By juxtaposing respiratory signals of the control group with the COVID-19 patients under identical breathing tests, we define a standard baseline. Such a baseline helps in discerning the impacts of COVID-19 on respiratory patterns and sets a clear standard for comparing data from the COVID-19 patients.

### 3.2. Feature Extraction

In this segment, we delve into a detailed feature extraction analysis based on the collated respiratory signal data. The aim is to underline the pronounced differences in respiratory patterns between healthy subjects and COVID-19 patients. **Fig. 2** sketches out the peak detection and power spectral density (PSD) analyses of respiratory signals for both cohorts across the three respiratory tests: normal breathing, breath holding, and deep breathing.

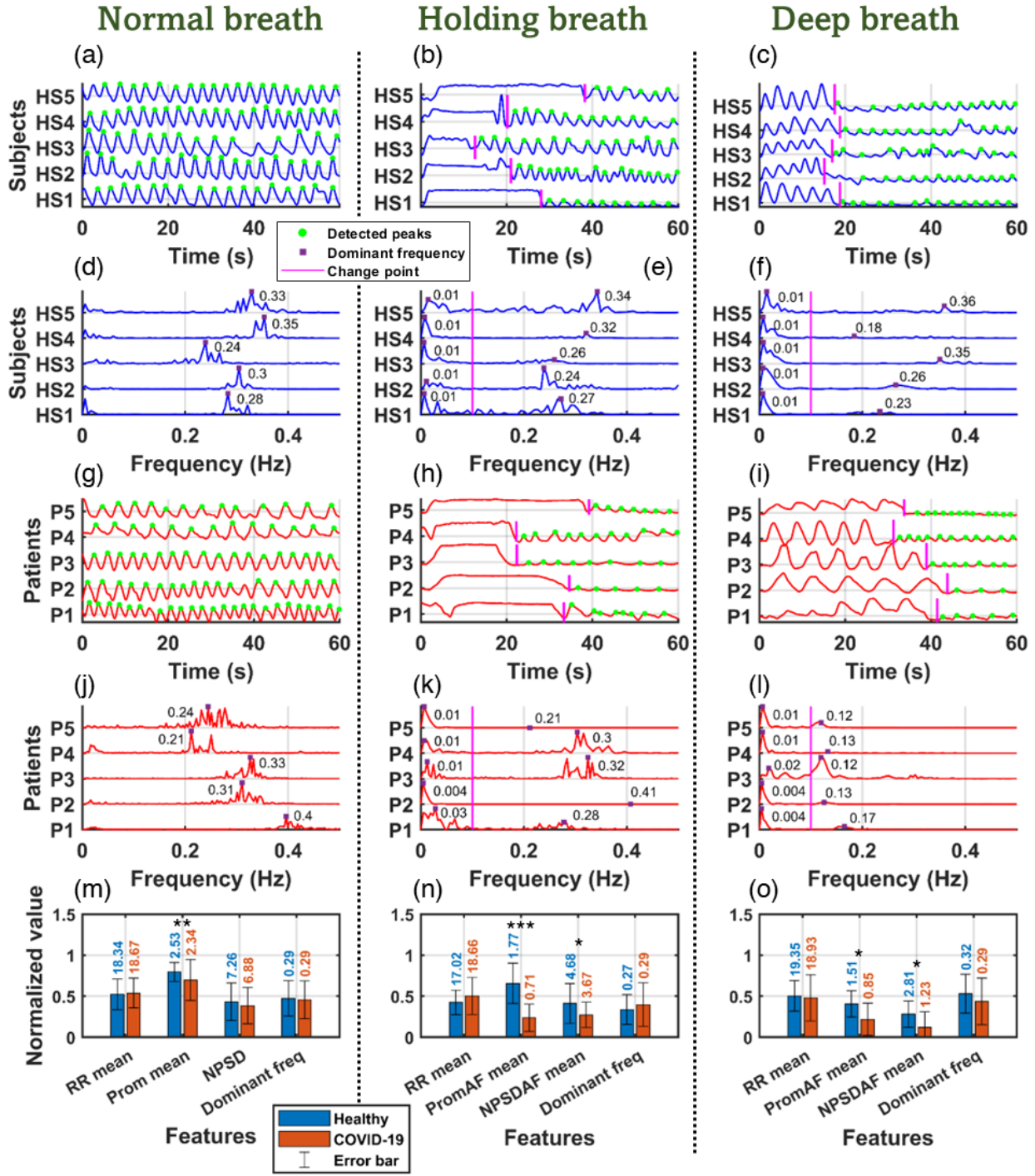

**Fig 2**. Analysis and feature extraction of respiratory signals for both healthy and COVID-19 subjects. (a), (b), (c), (d), (e), and (f) portray peak detection and power spectral density (PSD) analysis of respiratory signals for healthy participants during the three breathing tests: normal

breath, breath-hold, and deep breath. Similarly, (g), (h), (i), (j), (k), and (l) show peak detection and power spectral density analysis for COVID-19 patients following the same breath test protocols. (m), (n), (o) are bar charts comparing four representative extracted time-domain and frequency-domain features (respiration rate (RR) mean, prominence (Prom) mean, normalized power spectral density (NPSD), dominant frequency (freq)) between healthy subjects and COVID-19 patients. Significant variances in respiratory features between the two groups were validated using the two-sample t-test with a confidence level $\alpha = 0.05$ and significance codes: $p < 0.001$ '***', $p < 0.01$ '**', $p < 0.05$ '*'. HS stands for healthy subject, while P denotes COVID-19.

Upon visually examining the data, distinct variations in the peak patterns and their related spectral contents between healthy and affected individuals are evident (**Figs. 2a-2l**). Delving deeper into specific features, we identified and recorded four representative features from both the time and frequency domains. These were: the mean respiration rate (RR), mean prominence (Prom), normalized power spectral density (NPSD), and dominant frequency (Freq). We then compared the mean values of these features for both groups across the three different breathing conditions. The bar charts presented in **Figs. 2m-2o** reveal pronounced disparities in these features when comparing healthy and COVID-19 subjects, suggesting that the disease markedly influences respiratory dynamics.

To statistically evaluate these observations, we conducted a two-sample t-test, contrasting the respiratory features (RR, Prom, NPSD, Freq) of the healthy group against the COVID-19 group over three distinct breathing tests: normal, hold, and deep breathing, setting the significance level at $\alpha = 0.05$. For the normal breathing test, our findings indicated a significant variation in the prominence feature (Prom) between the groups, with a p-value of 0.0076. However, for other metrics like the respiration rate (RR), normalized power spectral density (NPSD), and dominant frequency (Freq), the variations were not deemed statistically significant ($p > 0.05$). When examining the breath-holding condition, we noted marked discrepancies in the Prom ($p < 0.001$) and NPSD ($p = 0.0107$) attributes between the two groups. Yet, no significant variations were discerned for the RR and Freq metrics ($p > 0.05$). As for the deep breathing test, significant variations were identified in the Prom ($p = 0.0123$) and NPSD ($p = 0.0265$) features. Once again, the RR and Freq metrics did not showcase statistically significant differences ($p > 0.05$). These outcomes suggest that our designated features, especially Prom and NPSD, exhibit notable distinctions between healthy participants and those diagnosed with COVID-19 across varied breathing conditions. Following this, we utilized RQA—a sophisticated non-linear analytical method—to encapsulate the dynamic traits of the respiratory signals, as depicted in **Fig. 3**.

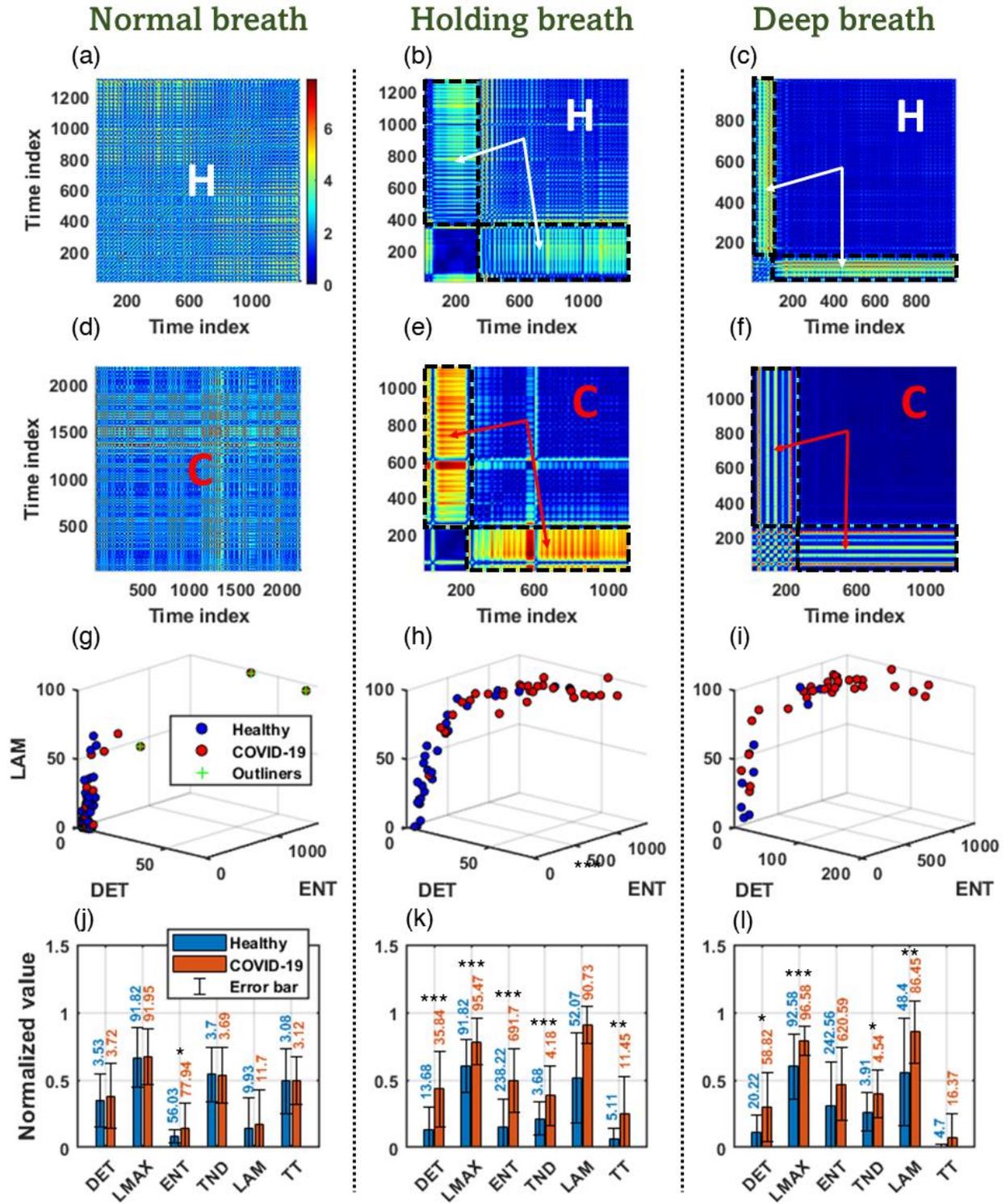

**Fig 3**. Advanced analysis and feature extraction of respiratory signals for both healthy subjects and COVID-19 patients using RQA. (a), (b), and (c) show the recurrence analysis plot for healthy

subjects used to quantify the recurrence of their respiratory signal dynamics observed from the three respiratory tests (normal, hold, and deep breathing). (d), (e), and (f) display the recurrence analysis plot for COVID-19 patients followed the same breath tests, with the black rectangular boxes indicating the "fingerprint" regions distinguishing the respiratory patterns and dynamics between healthy and COVID-19 subjects. (g), (h), and (i) are 3-D scatter plots of the three most distinctive features, namely DET, ENT, and LAM, for the three breathing tests. (j), (k), and (l) are bar charts comparing six RQA features of healthy and COVID-19 subjects with error bars. Significant variances were evaluated using the two-sample t-test with a confidence level $\alpha = 0.05$ and significance codes: $p < 0.001$ '***', $p < 0.01$ '**', $p < 0.05$ '*'. Note that the labels of *y*-axis for holding breath and deep breath are the same for normal breath and are omitted for clarity.

The RQA results, as presented in **Fig. 3,** emphasize distinctive recurrence plots for the two groups under three breathing conditions (**Figs. 3a-3f**). These plots vividly display specific areas, termed "fingerprint" regions, which uniquely characterize the intricate respiratory dynamics seen in COVID-19 patients, setting them apart from healthy individuals. These regions are marked by three pivotal RQA metrics: determinism (DET), entropy (ENT), and laminarity (LAM). Viewing the 3D scatter plots of these features for both groups (**Figs. 3g-3i**) shows well-separated clusters, especially for breath-holding and deep breathing. This suggests that these RQA metrics possess potent discriminatory capacity. Nevertheless, during normal respiratory tests, there is a noticeable overlap between data points from both healthy and COVID-19 subjects. Such overlap during standard breathing might indicate that the respiratory dynamics' alterations due to COVID-19 are not as discernible in this condition. While the disease indeed affects respiratory functionality, these changes might not be substantial enough to significantly influence standard relaxed breathing, preventing distinct clustering in the feature space. In addition, comparing all six RQA metrics between the two groups through bar charts (**Figs. 3j-3l**) confirmed these observations. Significant differences were found using two-sample t-tests with a confidence level of $\alpha = 0.05$, reinforcing the reliability of the selected RQA metrics.

Using the two-sample t-test, we further assessed the differences in RQA features (DET, LMAX, ENT, TND, LAM, TT) between the two groups across the three breathing scenarios. For regular breathing, only the entropy (ENT) feature displayed a notable difference between the groups (p = 0.0126). Other features did not show statistically significant variations (p > 0.05). During breath-holding, all RQA features presented marked differences between the groups, especially features such as DET (p = 1.12e-05) and ENT (p = 3.85e-07), underscoring that this breathing condition amplifies the disparities in respiratory dynamics for COVID-19 patients. As for deep breathing, significant variances were observed in DET (p = 0.0216), LMAX (p = 0.000659), TND (p = 0.0189), and LAM (p = 0.00354) metrics. However, ENT and TT did not demonstrate statistically significant discrepancies (p > 0.05). Collectively, these findings suggest that RQA metrics, particularly during breath-holding, effectively distinguish between the respiratory patterns of healthy individuals and those with COVID-19. This insight affirms that our selected features convey crucial and distinctive data regarding the respiratory behaviors of COVID-19 patients in contrast to healthy counterparts. Yet, it is vital to recognize that not all metrics present significant distinctions across each breathing condition, reflecting the intricate influence of COVID-19 on respiratory dynamics. Hence, quantifying these features' statistical significance becomes essential for the judicious choice of training features in ML models.

### 3.3. Feature selection and machine learning models

Following the extraction and detailed analysis of the respiratory signal features, our next step involved narrowing down to the most pertinent features through the feature selection phase. This helped us focus on the salient attributes and effectively reduce the dimensionality of our dataset. The visualization of this process is depicted in **Fig. 4**, where the outcomes for each of the three respiratory tests - normal, hold, and deep breathing - are presented.

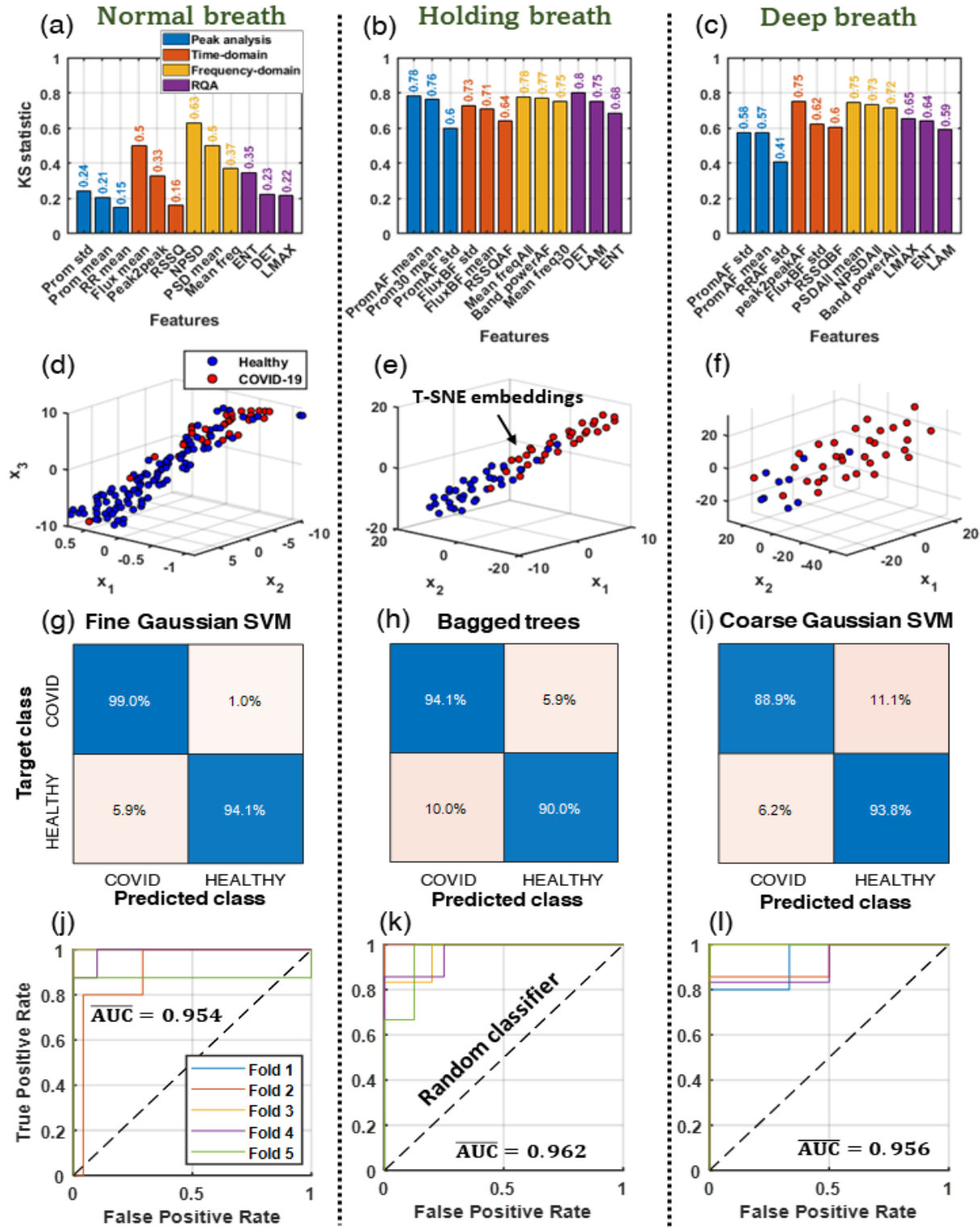

**Fig 4**. Feature selection, visualization, and performance evaluation of the ML model for COVID-19 diagnosis. (a), (b), and (c) display KS statistics for feature selection (ranking) across four feature

groups: time-domain, frequency-domain, peak analysis, and RQA for three respiratory tests (normal, hold, and deep breathing). (d), (e), and (f) show manifold learning using the t-SNE method for visualizing a high-dimensional feature space, highlighting the separation between healthy and COVID-19 subjects in 3-D space for the three respiratory tests. (g), (h), and (i) display the confusion matrix of the ML model demonstrating the accuracy in classifying COVID-19 cases and healthy subjects using 3 different breathing tests. (j), (k), and (l) show ROC curve plots of 5-fold cross-validation step indicating the model's robust performance and diagnostic capability, the average area under the curve $\overline{AUC}$ was reported.

We utilized the Kolmogorov-Smirnov (KS) statistics (**Figs. 4a-c**) to prioritize features from the four feature groups we examined: time-domain, frequency-domain, peak analysis, and RQA. For normal breathing, the most salient features across these groups were Flux mean, NPSD, Prom std, and ENT. During the breath-holding test, Flux BF std, Mean freqAll, Prom AF mean, and ENT stood out as the most discerning features. For deep breathing, peak2peakAF, PSDAll mean, Prom AF std, and LMAX were identified as the most indicative. These insights guided our focus on features with a marked ability to differentiate between healthy individuals and those with COVID-19, echoing our preliminary feature extraction and analysis.

The complex feature space was visualized via the t-SNE method in manifold learning (**Figs. 4d-f**). These 3D plots emphasize the distinction between healthy and COVID-19 subjects within the transformed feature domain, underlining the discriminating prowess of the chosen features. Post feature selection, we trained and assessed ML models for their efficacy in distinguishing between COVID-19 cases and healthy controls. The resultant confusion matrices (**Figs. 4g-i**) offer a snapshot of each model's classification capabilities, with notable accuracy in differentiating the two cohorts. These matrices supply a detailed breakdown of true positive, false positive, true negative, and false negative rates—critical metrics for gauging model performance. To further affirm the resilience and diagnostic potency of our ML models, we employed a 5-fold cross-validation and plotted the receiver operating characteristic (ROC) curves (**Figs. 4j-l**). The area under the curve (AUC) serves as a quantitative testament to each model's classification acumen. Notably, during normal breathing, the Fine Gaussian SVM model [42] emerged as the top performer, registering a sensitivity of 99%, a specificity of 94.1%, and an average ROC curve area ($\overline{AUC}$) of 0.954 across the 5-fold cross-validation. In the breath-holding scenario, the Bagged Trees model [43] stood out, albeit with a slightly reduced sensitivity of 94.1% and a specificity of 90%; its $\overline{AUC}$ value was 0.962. As for the deep breathing test, the Coarse Gaussian SVM model [42] yielded the prime outcome, with a sensitivity of 99%, specificity of 94.1%, and an $\overline{AUC}$ value of 0.956. These evaluations, spanning multiple respiratory tests and ML models, bolster our confidence in our feature selection's rigor and the ensuing models' aptitude in discerning healthy individuals from COVID-19 patients.

### 3.4. Causal analysis

To deepen our understanding of the connection between respiratory patterns and the presence of COVID-19, we conducted a causal analysis. We employed a matching method that classified patients into "treated" (those diagnosed with COVID-19) and "control" (healthy individuals) categories. Age, gender, and body mass index (BMI) were flagged as potential confounders that might simultaneously affect respiratory patterns and COVID-19 status. A comparison of unmatched and matched group results of healthy and COVID-19 subjects can be found in **Table 2**. The distribution dynamics of these confounders, pre and post the matching methodology, are showcased in **Fig. S3**, with a spotlight on crucial covariates like age, gender, and BMI. The standardized mean difference (SMD) in **Table 2** reveals a marked reduction in the mean variations related to age, gender distribution, and BMI between the two groups after matching. This suggests that the matching procedure was adept at formulating comparable groups.

**Table 2**. Pre-matching and post-matching comparison between healthy and COVID-19 groups.

| | UNMATCHED | | | MATCHED | | |
|---|---|---|---|---|---|---|
| | **Healthy** | **COVID-19** | **SMD**[*] | **Healthy** | **COVID-19** | **SMD** |
| **Participants**, *n* | 37 | 33 | | 33 | 33 | |
| **Age**, *mean (std)* | 40.91 (16.88) | 46.88 (13.64) | 0.398 | 46.26 (10.84) | 46.88 (13.64) | 0.034 |
| **Male** *n (%)* | 13 (35.14) | 14 (42.42) | 0.262 | 16 (48.48) | 14 (42.42) | < 0.001 |
| **BMI**, *mean (std)* | 22.03 (3.32) | 22.93 (2.99) | 0.293 | 23.47 (3.07) | 22.93 (2.99) | 0.031 |

*SMD = Standardized Mean Difference

The implications of COVID-19 on respiratory patterns, as delineated by our causal analysis, are depicted in **Fig. 5**. Specifically, **Figs. 5a-b** display the normalized values of six handpicked respiratory features for both healthy and COVID-19 groups after the matching method's application. These visualizations facilitate a deeper understanding of each group's feature distribution and highlight discernible shifts that might indicate COVID-19's effects. Transitioning to **Figs. 5c-d**, we shift focus from feature distribution to quantifying COVID-19's impact on these features, showcased through Average Causal Effect Percentages (ACEPs).

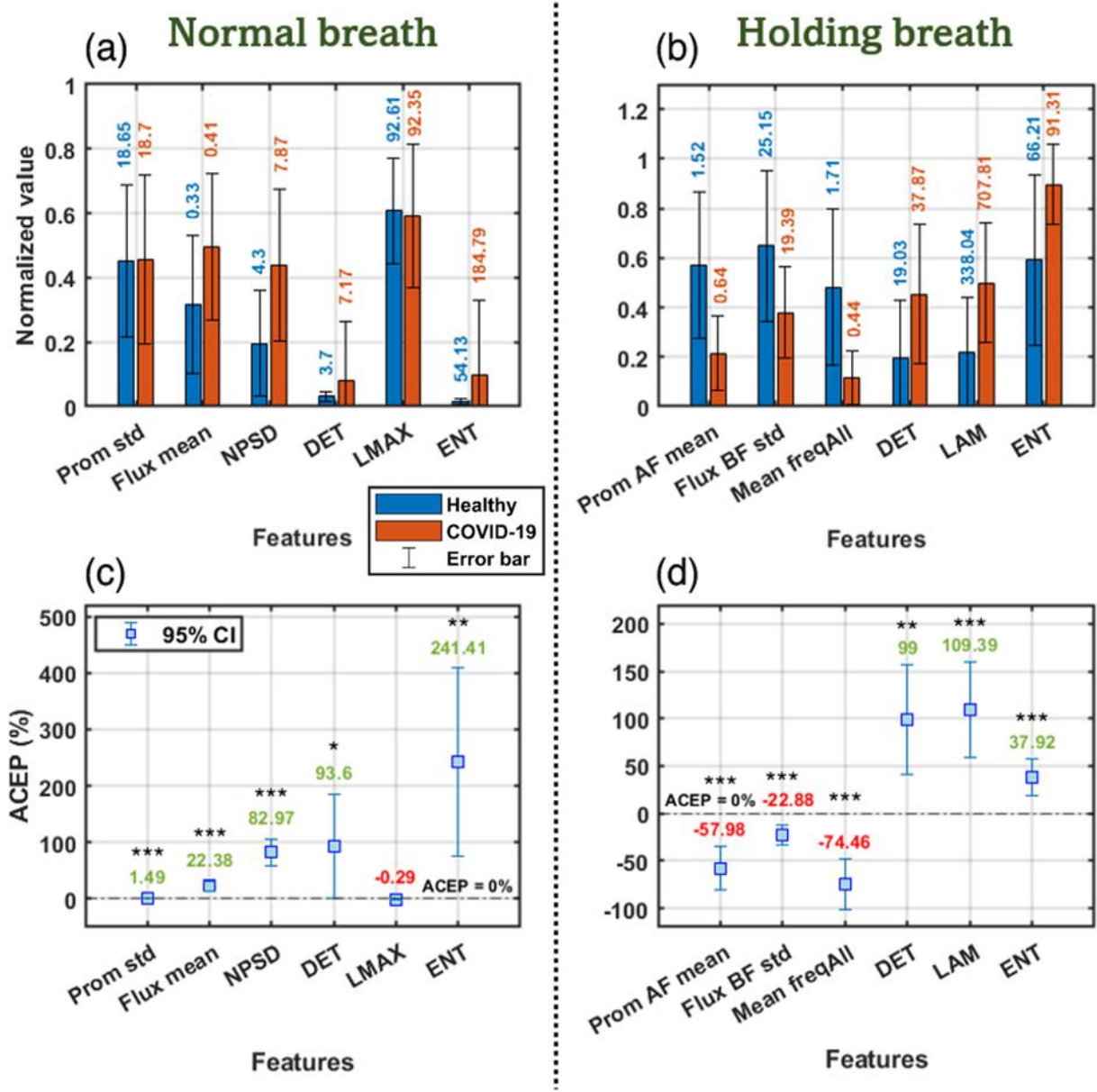

**Fig 5**. Causal analysis results illustrating the impact of COVID-19 on respiratory patterns. (a) and (b) show the normalized values of the six selected respiratory features for the healthy and COVID-19 groups after applying the matching method. Panel (a) corresponds to normal breathing, while panel (b) corresponds to the breath-holding test. (c) and (d) present the average causal effect percentage (ACEP) values for the six features of the normal breathing and breath-holding tests, respectively. ACEP was calculated by dividing the average causal effect (ACE) by the mean value

of the features of the healthy individuals, thereby expressing the causal effect in percentages relative to the healthy respiratory pattern. The significance of the causal effects was evaluated using the paired t-test with a confidence level $\alpha = 0.05$ and significance codes: $p < 0.001$ '***', $p < 0.01$ '**', $p < 0.05$ '*'.

The central goal of our causal analysis was to quantify COVID-19's influence on specific respiratory patterns. Moreover, we aimed to discern whether certain respiratory features could reliably differentiate between COVID-19 patients and healthy individuals using respiratory data. This exercise aimed to unearth potential pathophysiological changes in respiratory functions tied to COVID-19. As portrayed in **Fig. 5**, our data uncovers the profound alterations COVID-19 may introduce to respiratory patterns. This is evident from the significant levels associated with the chosen features. Yet, COVID-19's influence is multifaceted and can differ across various testing features and conditions, underscoring the disease's intricate impact on respiratory patterns. For instance, during normal breathing (see **Fig. 5c**), features like Prom std, Flux mean, NPSD, and ENT register extremely high significance ($p < 0.001$), reinforcing the palpable causal effects of COVID-19. Changes span from a minor -0.29% decrease in LMAX, which is not statistically significant ($p > 0.05$), to a substantial 241.41% increase in ENT ($p < 0.01$). This indicates notable alterations in certain respiratory patterns owing to COVID-19. Additionally, DET marked a significant causal effect with a 93.6% rise ($p < 0.05$), albeit with lesser confidence.

Findings from the breath-holding tests (**Fig. 5d**) echo these patterns, albeit with nuances in significance levels. All features, including Prom AF mean, Flux BF mean, Mean freqAll, DET, LAM, and ENT, showcase high significance ($p < 0.001$), although DET registers a slightly elevated p-value ($p < 0.01$). Impacts range from a marked -74.46% drop for Mean freqAll to a 109.39% surge for LAM, with DET also indicating a 99% rise. This data underpins the pronounced effects of COVID-19 on respiratory behavior. However, it is imperative to note the intricate nature of these changes—they can oscillate dramatically between features and test scenarios. This intricacy accentuates the importance of holistic and multivariate methodologies when studying COVID-19's respiratory ramifications. Comprehensive discussions on the pathophysiology associated with COVID-19-related shifts in the respiratory system, as gauged from these features, are provided in the **Supporting Information** and **Table S3.**

## 4. DISCUSSION

We showcased the potent capability of MRST when paired with ML-assisted analysis for real-time monitoring and diagnosis of COVID-19 and its variants. Most prior ML-based analyses for COVID-19 diagnostics utilized other data modalities, like images or electronic health records. To our knowledge, our study stands out as one of the pioneering endeavors harnessing respiratory signals for this end. Our research augments the existing literature in several pivotal aspects:

First, our study integrated an exhaustive set of features derived from respiratory signals. The feature extraction methodologies were conceived to encompass a wide array of characteristics intrinsic to these signals. The amalgamation of peak-derived, time-domain, frequency-domain and RQA features offers a nuanced insight into respiratory patterns. Additionally, we employed the change point detection technique on our respiratory data, underscoring its efficacy in breath-holding and deep breathing analyses. This technique discerns notable shifts in respiratory behaviors and allows the extraction of features both pre and post these change points. Scrutinizing these sharp changes unravels crucial insights into how the body reacts, facilitating the differentiation between COVID-19 afflicted patients and healthy subjects. In juxtaposition to many previous works that largely emphasized a singular feature domain, our expansive approach feeds our ML models with a more diverse and rich data spectrum, bolstering the classification between COVID-19 cases and healthy counterparts.

In a departure from numerous studies that solely employ a singular ML model, our approach harnessed an array of models from MATLAB's Classification Learner Toolbox. This approach enabled the juxtaposition of various models, streamlining the selection of the best-fitting model for our dataset. Notably, the Fine Gaussian SVM model, Bagged Trees model, and Coarse Gaussian SVM model emerged as top performers for normal, hold, and deep breathing tests respectively. A salient strength of our study was the consistent performance across diverse models under varying breathing scenarios. This consistency underscores the sturdiness of our chosen features across different physiological states and the models' adeptness in adjusting to diverse breathing nuances — a crucial facet given COVID-19's varied clinical presentations. Such versatility further amplifies diagnostic precision and mitigates the risk of erroneous outcomes. With commendable sensitivity and specificity, these models underscore their potential in providing accurate and dependable COVID-19 detection. Our methodology's stellar performance, even in the face of unseen data, testifies to the approach's resilience and broad applicability.

Furthermore, our causal analysis elucidated the pronounced effects of COVID-19 on distinct respiratory features. The derived ACEP values spotlight potential deviations in respiratory patterns attributed to COVID-19, manifesting as either considerable increases or decreases, contingent on the specific feature and test type. Most of these variations bore statistical significance, underscoring COVID-19's profound causal influence on respiratory behaviors. Insights harvested from this analysis are pivotal, shedding light on the respiratory nuances in COVID-19 patients. This understanding could potentially catalyze the creation of superior diagnostic tools and therapeutic modalities.

When juxtaposed with existing literature methodologies, our combination of MRST and ML-assisted analysis boasts several distinctive merits. Primarily, our method facilitates real-time monitoring of COVID-19, contrasting many diagnostic tools that necessitate extended analysis or multiple steps [44, 45]. This immediacy is crucial in clinical scenarios where swift diagnosis can profoundly impact patient outcomes [46]. While we respect the significant strides achieved by preceding methods, especially those employing imaging or electronic health records [14-18], our technique focuses on the nuances of respiratory signals, setting it apart. Leveraging a diverse array of respiratory features, our ML models are fed with a richer dataset compared to many traditional techniques [47, 48], which often limit their feature sets. This holistic input enhances our model's precision in differentiating COVID-19 patients from healthy individuals across varied breathing patterns. Additionally, the non-contact nature of our magnetic respiratory sensor ensures a non-intrusive patient experience. Many diagnostics in literature, especially those demanding direct contact or invasive procedures [49, 50], can distress patients. In contrast, our method seeks to maximize patient comfort, thus promoting greater adherence to testing. It is also worth noting that the comfort afforded by our approach may indirectly bolster its accuracy, a benefit not always quantified in existing literature.

However, our study is not devoid of constraints. Primarily, the data was sourced from a niche demographic and geographical sector with a limited sample size, potentially constraining the model's broader applicability. Expanding research to encompass more diverse patient groups is imperative for a comprehensive validation of our findings. Secondly, while our model adeptly classifies COVID-19 cases, its efficacy in differentiating COVID-19 from other respiratory ailments remains uncharted territory. Given the symptom overlap between these diseases and COVID-19, there's potential for misdiagnoses. It is crucial that future work explores our model's discernment capabilities among these respiratory conditions. Thirdly, our magnetic respiratory sensor, while pioneering in offering non-contact diagnostics, might introduce noise into the respiratory data, potentially impacting feature extraction and ML processes. Enhancements in signal acquisition and refining feature algorithms are paramount. Moreover, our analysis was rooted in a singular data collection point per participant. Factoring in longitudinal data could enhance our models by acknowledging individual variations and respiratory pattern shifts over time. Future endeavors will seek broader data collection, probe our model's differential diagnosis prowess, refine signal collection, and weave in longitudinal data for a comprehensive analysis.

The efficacy of our models lays the groundwork for swift triaging of potential COVID-19 cases across varied environments. Incorporating these models into real-time monitoring systems empowers healthcare professionals to incessantly scrutinize respiratory data in high-risk zones like intensive care units. Such systems can enable swift detection, alert healthcare units promptly, and pave the way for early interventions, potentially curtailing disease progression and transmission. For expedited diagnoses, our models stand as pivotal screening tools in bustling healthcare setups, transit points, or massive public gatherings. They can pinpoint individuals necessitating comprehensive testing, alleviating medical resource strain and accelerating the diagnostic process amid potential outbreaks. In a wider perspective, our insights also hint at applications for managing respiratory maladies like COPD and asthma. Our method's ability to extract detailed respiratory signal features can unveil nuanced respiratory patterns, heralding a new era in disease monitoring, personalized treatments, and targeted interventions. Embedding our technology into devices like wearables, smartphone applications, or smart home systems could transform everyday gadgets into potent, discreet health monitors. Such platforms could bestow both individuals and healthcare providers with constant, real-time insights into respiratory patterns, prompt timely responses for vulnerable groups, and potentially reshape the global approach to respiratory health management.

## 5. CONCLUSION

We have presented a pioneering approach to the real-time monitoring and diagnosis of COVID-19 and its variants by harnessing the combined strengths of MRST and ML. Our research underscores the viability of respiratory signal features in distinguishing COVID-19 patients from healthy individuals, marking a significant contribution to the global initiatives battling the pandemic. The efficacy of this study stems from our meticulous selection of pertinent respiratory features, adept application of sophisticated ML algorithms, and stringent validation protocols. Crucially, our method not only stands as a promising independent diagnostic tool but also augments the reliability and speed of current testing methodologies. The capability for non-invasive and remote monitoring of respiratory signals holds the potential to revolutionize COVID-19 screening techniques, particularly in settings with limited resources or scenarios demanding swift outcomes. Our work enriches the burgeoning domain of digital health solutions, epitomizing the immense promise of interdisciplinary endeavors that draw from health science, engineering, and data science to address pressing global health challenges. This research will undoubtedly catalyze further explorations aimed at honing and broadening our techniques, thereby enhancing the future landscape of healthcare monitoring and diagnostic tools.

### ACKNOWLEDGEMENTS

Research at the University of South Florida was partially supported by the USF COVID-19 Research Foundation under grant number 125300. We thank the research assistants from Vietnam for their help in collecting the data. The authors thank Dr. Tatiana Eggers and Mr. Noah Schulz for editing the English of the manuscript and for their useful comments.

**DECLARATION OF COMPETING INTERESTS**

The authors declare that they have no known competing financial interests or personal relationships that could have appeared to influence the work reported in this paper.

# Supplementary Information

**Real-Time Magnetic Monitoring and Diagnosis of COVID-19 via Machine Learning**

Dang Nguyen[1,2#], Phat K. Huynh[3#], Vinh Duc An Bui[4,*], Kee Young Hwang[1], Nityanand Jain[5], Chau Nguyen[6], Le Huu Nhat Minh[7], Le Van Truong[8], Xuan Thanh Nguyen[9], Dinh Hoang Nguyen[10], Le Tien Dung[11], Trung Q. Le[2,3,*], and Manh-Huong Phan[1,*]

[1] Department of Physics, University of South Florida, Tampa, FL 33620, USA

[2] Department of Medical Engineering, University of South Florida, Tampa, FL 33620, USA

[3] Department of Industrial and Management Systems Engineering, University of South Florida, Tampa, FL 33620, USA

[4] Department of Thoracic and Cardiovascular Surgery, Hue Central Hospital, Hue City, Vietnam

[5] Faculty of Medicine, Riga Stradiņš University, 16 Dzirciema street, Riga, LV-1007, Latvia

[6] Vietnam National University, Ho Chi Minh City, Vietnam

[7] Emergency Department, University Medical Center, Ho Chi Minh City, Vietnam

[8] Traditional Medicine Hospital, Ministry of Public Security, Hanoi 10000, Vietnam

[9] Department of Abdominal Emergency and Pediatric Surgery, Hue Central Hospital, Hue City, Vietnam

[10] Department of Adult Cardiovascular Surgery, University Medical Center Ho Chi Minh City, Ho Chi Minh City, Vietnam

[11] Department of Lung, University Medical Center Ho Chi Minh City, Ho Chi Minh City, Vietnam

[#] Co-first authors

[*] Co-corresponding authors: phanm@usf.edu (M.H.P.), tqle@usf.edu (T.Q.L.), buiducanvinh@gmail.com (V.D.A.B.)

# 1. The working principles of the Hall effect sensor and the magnetic respiratory monitoring system

Magnetic respiratory sensors detect changes in magnetic flux. By placing a small magnet on a subject's chest, the magnetic flux of the atmosphere is altered due to the chest's movement during breathing (**Fig. S1**). In contrast to piezoelectric or impedance pneumogram sensors, which directly measure chest movement and volume through contact, magnetic respiratory sensors are contactless. Their enhanced sensitivity enables the differentiation of distinct breathing patterns [1].

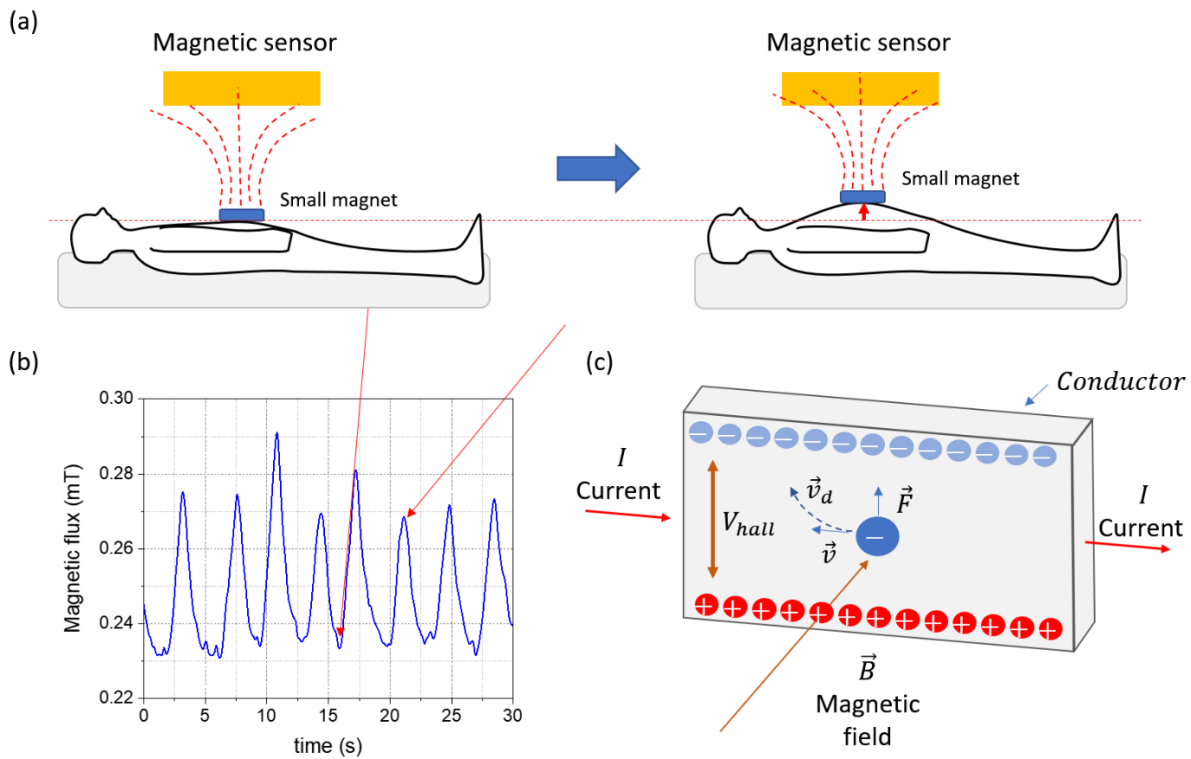

**Figure S1.** (a) A small permanent magnet placed on the subject's chest will change the magnetic flux on the magnetic sensor due to respiratory movements; (b) The breath pattern shows the peak corresponding to inhalation and the valley corresponding to exhalation; (c) the basic concept of the Hall effect and the magnetic sensor based on this effect.

Hall sensors, well known for their magnetic sensing capabilities, seamlessly integrate into electronic gadgets. Their versatility and CMOS compatibility render them valuable in numerous domains, spanning from automotive and consumer electronics to industrial control systems and robotics [2]. These sensors' sensitivity is adaptable. While they can measure high magnetic fields—like those in fusion applications where field strengths might surpass 20 Tesla (T)—they can also discern minute magnetic fields below 1 milli Tesla (mT) [3]. Contemporary Hall elements, constructed from ultra-thin semiconductor slices, ensure that these magnetic sensors are both cost-effective and extraordinarily sensitive (**Fig. S2**) [4]. The Hall effect gives rise to a potential difference across a conductor when a magnetic field operates perpendicularly to an induced current (**Fig. S1c**). When a magnetic field is applied, carriers will experience the Lorentz force which separates the carriers to one side of the conductor. Equilibrium is reached when

$$eE = ev_d B$$

where $e$ is the magnitude of the electron charge, $v_d$ is the drift speed, $E$ is the electric field by Lorentz force and $B$ is the perpendicular external magnetic field. $v_d$ can be written as

$$v_d = \frac{E}{B}$$

If the current due to Lorentz force on the material is $I$,

$$I = nev_d A$$

where $n$ is the number of charge and $A$ is the cross section, $I$ can be written in the other form using $v_d$,

$$I = ne\left(\frac{E}{B}\right)A$$

Electric potential between the edge of the material due to the Hall effect is

$$V = El$$

where $l$ is the distance between the edge. Finally, we can obtain

$$V = \frac{IBl}{neA} = Blv_d$$

as the Hall voltage [2]. As the Hall effect sensor quantifies a magnetic field's magnitude, the output voltage is directly proportional to the magnetic field strength passing through it (**Fig. S1c**). The contemporary Hall effect sensor incorporates a single integrated circuit, housing one conventional planar Hall element and dual vertical Hall element sets. A rudimentary model for the vertical Hall element resembles N-type silicon plates inserted vertically into a P-type substrate (**Fig. S2a**). An external magnetic field induces an opposing current in the plate's center when current flows from the plate's center to its endpoints. By organizing these elements appropriately, the sensor can gauge a 3-axis magnetic field with heightened sensitivity [4]. An exemplary Hall sensor adept at precisely measuring respiratory signals is the commercially available THM1176-LF from MetroLab. Operating within an 8mT magnetic field range, it offers ±20 µT accuracy and ±2 µT resolution within an active 6×3.4×3 mm volume, making it ideally suited for real-time respiratory pattern monitoring with high precision [4].

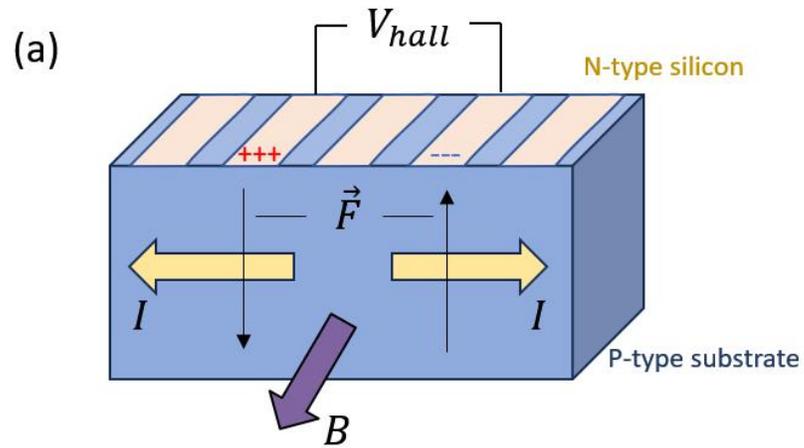

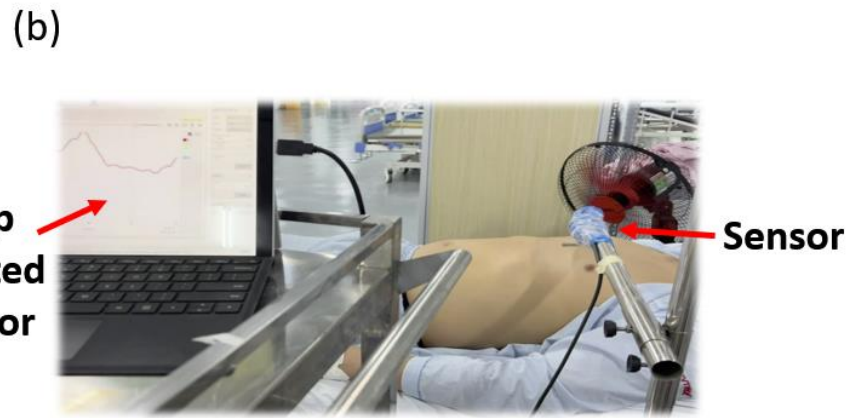

**Figure S2.** Hall Effect sensor configuration and measurement setup. Panel (a) depicts a simple model of the vertical Hall element. Panel (b) showcases the practical setup used for respiratory monitoring in the hospital. It depicts the Hall effect sensor connected to a tablet PC via a data acquisition interface. The sensor, placed close to the subject's chest or abdomen, measures changes in the magnetic field associated with respiration, and the tablet PC records and displays this data in real time.

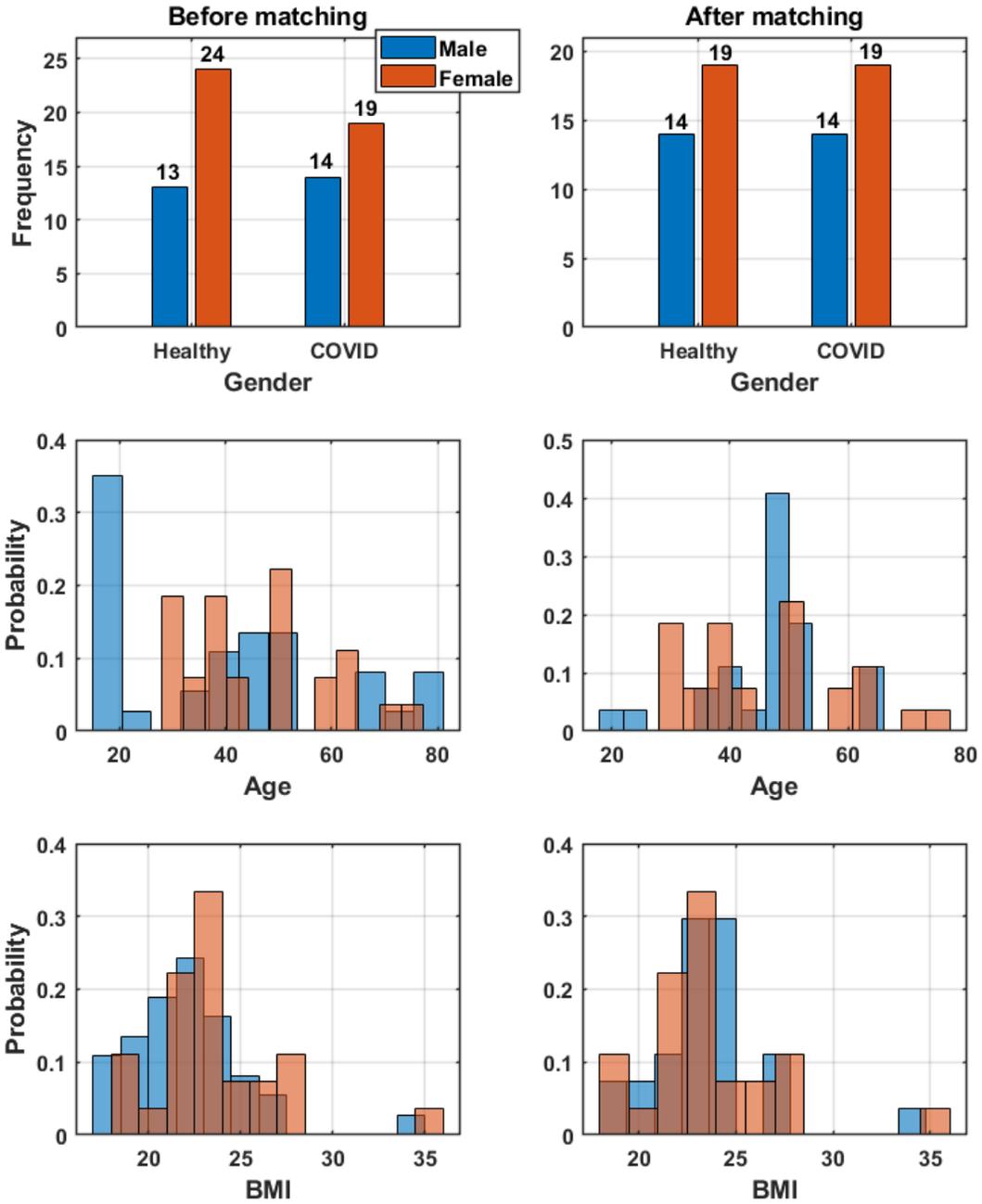

**Figure S3**. Pre- and post-matching/weighting comparison. This demonstrates the effectiveness of the causal analysis approach in balancing the observed covariates between the two groups.

**Table S1**. Summary of the extracted respiratory signal features. The features are divided into four main categories: peak-derived, time-domain, frequency-domain, and recurrence quantification analysis (RQA).

| Feature Category | Extracted Features | Mathematical formulation | Description |
|---|---|---|---|
| **Peak-derived Features** | Peak prominence mean $\bar{P}$ | $$\bar{P} = \frac{1}{N}\sum_{i=1}^{N} P_i, \qquad P_i = \max(Y_i) - \min(Y_i)$$ | The measure of how much a peak stands out due to its intrinsic height and its relative location. The mean and standard deviation (std) capture the central tendency and dispersion of the prominences in the respiratory signal. |
| | Peak prominence std $s_P$ | $$s_P = \sqrt{\frac{1}{N-1}\sum_{i=1}^{N}(P_i - \bar{P})^2}$$ where $Y_i$ is the $i^{th}$ peak, $N$ is the number of peaks, $P_i$ represents the difference between the peak (maximum) $\max(Y_i)$ and trough (minimum) $\min(Y_i)$ of the signal. | |
| | Peak width mean $\bar{W}$ | $$\bar{W} = \frac{1}{N}\sum_{i=1}^{N} W_i,$$ | The width of the peaks at their half-prominence height. The mean and std of the peak widths provide insights into the typical width and variability of the breath cycles. |
| | Peak width std $s_W$ | $$s_W = \sqrt{\frac{1}{N-1}\sum_{i=1}^{N}(W_i - \bar{W})^2}$$ where $W_i$ is the width of the $i^{th}$ peak at their half-prominence height. | |
| | Respiration rate mean $\overline{RR}$ | $$\overline{RR} = \frac{1}{N-1}\sum_{i=1}^{N-1} R_i$$ | The mean and standard deviation of the respiration rates provide a measure of average breathing rate and its variability. |
| | Respiration rate std $s_{RR}$ | $$s_{RR} = \sqrt{\frac{1}{N-2}\sum_{i=1}^{N}(R_i - \overline{RR})^2}$$ where $R_i$ is the respiration rate for the $i^{th}$ peak-to-peak interval | |
| **Time-Domain Features** | Maximum-to-minimum | $$mmd = \max(X_i) - \min(X_i)$$ | The difference between the maximum and |
| | difference $mmd$ | where $\max(X_i)$ and $\min(X_i)$ denote the maximum and minimum values in the respiratory signal $\boldsymbol{X}$. | minimum values in the respiratory signal. |
| | Root-sum-of-squares level $rssq$ | $$rssq = \sqrt{\sum_{i=1}^{N_{data}} |X_i|^2}$$ where $N_{data}$ is the number of samples and $X_i$ is the $i^{th}$ sample of the signal $\boldsymbol{X}$. | Overall measure of the signal's magnitude. |
| | Flux amplitude mean $\overline{Flux}$ | $$\overline{Flux} = \frac{1}{N_{data}} \sqrt{\sum_{i=1}^{N_{data}} X_i}$$ | The mean and std of flux amplitude captures the average and variability of the flux amplitude changes. |
| | Flux amplitude std $s_{Flux}$ | $$s_{Flux} = \sqrt{\frac{1}{N_{data}-1} \sum_{i=1}^{N} \left(X_i - \overline{Flux}\right)^2}$$ | |
| **Frequency-Domain Features** | Band power $BP$ | $$BP = \int_{f_1}^{f_2} |\widehat{\boldsymbol{X}}(f)|^2 df$$ where $\widehat{\boldsymbol{X}}(f)$ is the Fourier Transform of the signal $\boldsymbol{X}(t)$ and $[f_1, f_2]$ is the frequency band. | This measures the power of the signal within a specific frequency band. |
| | Power spectral density (PSD) mean $\overline{PSD}$ | $$\overline{PSD} = \int_{f \in \mathcal{F}} |\widehat{\boldsymbol{X}}(f)|^2 df$$ | PSD shows how the power of a signal is distributed with frequency. $\overline{PSD}$ provides an average power value across frequencies, while the normalized PSD scales the PSD to a standardized range. |
| | Normalized PSD $NPSD$ | $$NPSD = \frac{1}{T} \int_{f \in \mathcal{F}} |\widehat{\boldsymbol{X}}(f)|^2 df$$ where $T$ represents the total time duration of the signal $\boldsymbol{X}$. | |
| | Mean frequency $f_{mean}$ | $$f_{mean} = \frac{\int_{f \in \mathcal{F}} f |\widehat{\boldsymbol{X}}(f)|^2 df}{\int_{f \in \mathcal{F}} |\widehat{\boldsymbol{X}}(f)|^2 df}$$ | This is the average frequency of the signal, weighted by the power at each frequency. |

| | | | |
|---|---|---|---|
| | Dominant frequency $f_{dom}$ | $$f_{dom} = \underset{f}{\mathrm{argmax}} \frac{\left|\hat{X}(f)\right|^2}{N \cdot fs}$$ where $fs$ is the sampling frequency. | This is the frequency with the maximum power in the PSD. |
| **Recurrence Quantification Analysis (RQA) Statistics** | Determinism $DET$ | $$DET = \frac{\sum_{l=l_{min}}^{N_p} \ell P(\ell)}{\sum_{l=1}^{N} \ell P(\ell)}$$ where $P(\ell)$ is the frequency distribution of the lengths $\ell$ of the diagonal lines which have at least a length of $\ell_{min}$, and $N_p$ is the total number of data points. | Proportion of recurrence points that form the diagonal line. |
| | Maximum line length $LMAX$ | $$LMAX = \max(\ell)$$ | The maximum length of the diagonal line in the recurrence plot. |
| | Entropy $ENT$ | $$ENT = -\sum_{l=l_{min}}^{N_p} p(\ell) \ln p(\ell)$$ $$p(\ell) = \frac{P(\ell)}{\sum_{l=l_{min}}^{N_p} P(\ell)}$$ | Measure of complexity or predictability of the time series. |
| | Trend $TND$ | $$TND = \frac{\sum_{i=1}^{\tilde{N}} \left(i - \frac{\tilde{N}}{2}\right)(RR_i - \langle RR_i \rangle)}{\sum_{i=1}^{\tilde{N}} \left(i - \frac{\tilde{N}}{2}\right)^2}$$ $$RR_k = \frac{1}{N_p - k} \sum_{k=j-i}^{N_p-k} R(i,j)$$ where $R(i,j)$ is the recurrence indicator function for the $i^{th}$ and $j^{th}$ points, $\langle \cdot \rangle$ is the average value, and $\tilde{N} < N$. | Measure of the monotonicity of the time series. |
| | Laminarity $LAM$ | $$DET = \frac{\sum_{v=v_{min}}^{N_p} v P(v)}{\sum_{v=1}^{N} v P(v)}$$ where $P(v)$ is the frequency distribution of the lengths $v$ of the vertical lines which have at least a length of $v_{min}$. | Proportion of recurrence points that forms a vertical line |

| | Trapping time $TT$ | $$TT = \frac{\sum_{v=v_{min}}^{N_p} vP(v)}{\sum_{v=1}^{N} P(v)}$$ | Average time the system stays in a state before moving to another state. |
|---|---|---|---|

**Table S2**. Summary of the extracted respiratory signal features. The features are divided into four main categories: peak-derived, time-domain, frequency-domain, and recurrence quantification analysis (RQA).

| Breathing test | Feature Category | Abbreviation | Description |
|---|---|---|---|
| NORMAL BREATH | Peak-derived Features | Prom mean | Peak prominence mean $\bar{P}$ |
| | | Prom std | Peak prominence standard deviation $s_P$ |
| | | Width mean | Peak width mean $\bar{W}$ |
| | | Width std | Peak width standard deviation $s_W$ |
| | | RR mean | Respiration rate mean $\overline{RR}$ |
| | | RR std | Respiration rate standard deviation $s_{RR}$ |
| | Time-Domain Features | Flux mean | Flux amplitude mean $\overline{Flux}$ |
| | | Flux std | Flux amplitude std $s_{Flux}$ |
| | | Peak2peak | Maximum-to-minimum difference $mmd$ |
| | | RSSQ | Root-sum-of-squares level $rssq$ |
| | Frequency-Domain Features | Band power | Band power $BP$ |
| | | PSD mean | Power spectral density mean $\overline{PSD}$ |
| | | NPSD | Normalized PSD $NPSD$ |
| | | Mean freq | Mean frequency $f_{mean}$ |
| | | Dom freq | Dominant frequency $f_{dom}$ |
| HOLDING BREATH AND DEEP BREATH | Peak-derived Features | PromBF mean | $\bar{P}$ before the change point |
| | | PromBF std | $s_P$ before the change point |
| | | WidthBF mean | $\bar{W}$ before the change point |
| | | WidthBF std | $s_W$ before the change point |
| | | RRBF mean | $\overline{RR}$ before the change point |
| | | RRBF std | $s_{RR}$ before the change point |
| | | PromAF mean | $\bar{P}$ after the change point |
| | | PromAF std | $s_P$ after the change point |
| | | WidthAF mean | $\bar{W}$ after the change point |
| | | WidthAF std | $s_W$ after the change point |
| | | RRAF mean | $\overline{RR}$ after the change point |
| | | RRAF std | $s_{RR}$ after the change point |
| | | Prom30 mean | $\bar{P}$ after 30 seconds of the signal |
| | | Prom30 std | $s_P$ after 30 seconds of the signal |
| | | Width30 mean | $\overline{W}$ after 30 seconds of the signal |
| | | Width30 std | $s_W$ after 30 seconds of the signal |
| | | RR30 mean | $\overline{RR}$ after 30 seconds of the signal |
| | | RR30 std | $s_{RR}$ after 30 seconds of the signal |
| | **Time-Domain Features** | FluxBF mean | $\overline{Flux}$ before the change point |
| | | FluxBF std | $s_{Flux}$ before the change point |
| | | Peak2peakBF | $mmd$ before the change point |
| | | RSSQBF | $rssq$ before the change point |
| | | FluxAF mean | $\overline{Flux}$ after the change point |
| | | FluxAF std | $s_{Flux}$ after the change point |
| | | Peak2peakAF | $mmd$ after the change point |
| | | RSSQAF | $rssq$ after the change point |
| | | Flux30 mean | $\overline{Flux}$ after 30 seconds of the signal |
| | | Flux30 std | $s_{Flux}$ after 30 seconds of the signal |
| | | Peak2peak30 | $mmd$ after 30 seconds of the signal |
| | | RSSQ30 | $rssq$ after 30 seconds of the signal |
| | **Frequency-Domain Features** | Band powerBF | $BP$ before the change point |
| | | PSDBF mean | $\overline{PSD}$ before the change point |
| | | NPSDBF | $\overline{W}$ before the change point |
| | | Mean freqBF | $s_W$ before the change point |
| | | Dom freqBF | $\overline{RR}$ before the change point |
| | | Band powerAF | $BP$ before the change point |
| | | PSDAF mean | $\overline{PSD}$ before the change point |
| | | NPSDAF | $\overline{W}$ before the change point |
| | | Mean freqAF | $s_W$ before the change point |
| | | Dom freqAF | $\overline{RR}$ before the change point |
| | | Band power30 | $BP$ before the change point |
| | | PSD30 mean | $\overline{PSD}$ before the change point |
| | | NPSD30 | $\overline{W}$ before the change point |
| | | Mean freq30 | $s_W$ before the change point |
| | | Dom freq30 | $\overline{RR}$ before the change point |
| | | Band powerAll | $BP$ before the change point |
| | | PSDAll mean | $\overline{PSD}$ before the change point |
| | | NPSDAll | $\overline{W}$ before the change point |
| | | Mean freqAll | $s_W$ before the change point |
| | | Dom freqAll | $\overline{RR}$ before the change point |

**Table S3**. Pathophysiological interpretations of respiratory features in COVID-19 patients.

| Feature | Potential Pathophysiological Interpretation |
|---------|---------------------------------------------|
| Prom std | Variability in breathing amplitude, possibly reflecting irregular breathing patterns due to inflammation or compromised lung function in COVID-19 patients. |
| Flux mean | Altered airway resistance due to inflammation and narrowing of the airways, which may change the flow of air during breathing in COVID-19 patients. |
| NPSD | Variation in the rhythm or strength of breathing in COVID-19 patients, possibly as a response to hypoxia or other respiratory distress. |
| ENT | Increased respiratory complexity, suggesting irregular and non-linear breathing patterns in COVID-19 patients, possibly as a compensatory mechanism activated due to respiratory distress. |
| Prom AF mean | Difficulty in maintaining consistent breath holds in COVID-19 patients, possibly due to compromised lung function or discomfort. |
| Flux BF mean | Altered comfort or ability to hold breath in COVID-19 patients, potentially reflecting obstructed airways or impaired gas exchange. |
| Mean freq All | Slowed or labored breathing in COVID-19 patients, potentially indicating inflammation or hypoxia that affects the frequency of respiratory signals. |
| LAM | Periods of more regular and less chaotic respiratory system in COVID-19 patients, possibly indicative of altered lung mechanics or a compensatory deeper and more controlled breathing pattern due to impaired gas exchange. |
| DET | Increased determinism in the breathing pattern of COVID-19 patients, indicating a more repetitive and less variable breathing pattern, which might be a response to respiratory stress or hypoxia. |